\documentclass[runningheads]{llncs}

\usepackage{multirow}
\usepackage[inline]{enumitem}
\usepackage{xkeyval}
\usepackage{pax}
\usepackage{booktabs}
\usepackage{amsmath}
\usepackage{amssymb}
\usepackage{ctable}

\makeatletter
\newcommand{\printfnsymbol}[1]{%
  \textsuperscript{\@fnsymbol{#1}}%
}
\makeatother

\makeatletter
\define@key{PAX@Gin}{scale}{}
\makeatother

\begin{document}

\title{3D-PSRNet: Part Segmented 3D Point Cloud Reconstruction From a Single Image} 

\titlerunning{Part Segmented 3D Reconstruction}

\authorrunning{Mandikal, K L, Babu}

\author{Priyanka Mandikal\thanks{equal contribution} \and
Navaneet K L\printfnsymbol{1} \and R. Venkatesh Babu}

\institute{Indian Institute of Science, Bangalore, India
\email{priyanka.mandikal@gmail.com,\{navaneetl,venky\}@iisc.ac.in}\\
}

\maketitle

\begin{abstract}
We propose a mechanism to reconstruct part annotated 3D point clouds of objects given just a single input image. We demonstrate that jointly training for both reconstruction and segmentation leads to improved performance in both the tasks, when compared to training for each task individually. The key idea is to propagate information from each task so as to aid the other during the training procedure. Towards this end, we introduce a \textit{location-aware segmentation loss} in the training regime. We empirically show the effectiveness of the proposed loss in generating more faithful part reconstructions while also improving segmentation accuracy. We thoroughly evaluate the proposed approach on different object categories from the ShapeNet dataset to obtain improved results in reconstruction as well as segmentation. Codes are available at \url{https://github.com/val-iisc/3d-psrnet}.

\keywords{Point cloud, 3D reconstruction, 3D part segmentation}
\end{abstract}

\section{Introduction}

\begin{figure*}
\centering
\begin{center}
\includegraphics[width=\linewidth]{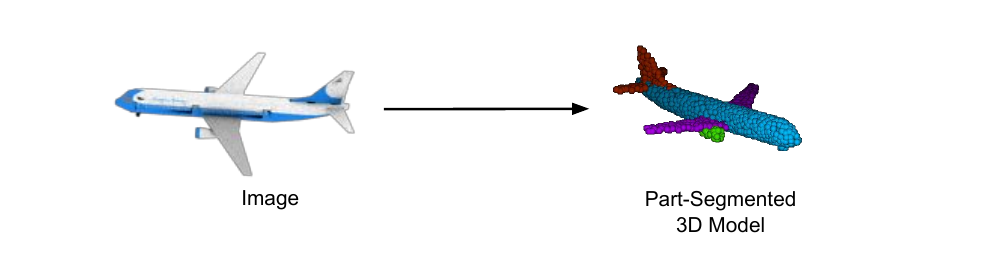}
\end{center}
\caption{Semantic point cloud reconstruction.}
\label{fig:intro}
\end{figure*}

Human object perception is based on semantic reasoning~\cite{koopman2017evolutionary}. When viewing the objects around us, we can not only mentally estimate their 3D shape from limited information, but we can also reason about object semantics. For instance, upon viewing the image of an airplane in Figure~\ref{fig:intro}, we might deduce that it contains four distinct parts - body, wings, tail, and turbine. Recognition of these parts further enhances our understanding of individual part geometries as well as the overall 3D structure of the airplane. This ability to perceive objects driven by semantics is important for our interaction with the world around us and the manipulation of objects within it.

In machine vision, the ability to infer the 3D structures from single-view images has far-reaching applications in the field of robotics and perception. Semantic understanding of the perceived 3D object is particularly advantageous in tasks such as robot grasping, object manipulation, etc.

Deep neural networks have been successfully employed for tackling the problem of 3D reconstruction. Most of the existing literature propose techniques for predicting the voxelized representation format. However, this representation has a number of drawbacks. First, it suffers from sparsity of information. All the information that is needed to perceive the 3D structure is provided by the surface voxels, while the voxels within the volume increase the representation space with minimal addition of  information. Second, the neural network architectures required for processing and predicting 3D voxel maps make use of 3D CNNs, which are computationally heavy and lead to considerable overhead during training and inference. For these reasons, there have been concerted efforts to explore representations that involve reduced computational complexity compared to voxel formats. Very recently, there have been works focusing on designing neural network architectures and loss formulations to process and predict 3D point clouds ~\cite{qi2017pointnet,qi2017pointnet++,fan2017point,su2018splatnet,li2018pointcnn}. Since point clouds consist of points being sampled uniformly on the object's surface, they are able to encode maximal information about the object's 3D characteristics. The information-rich encoding and compute-friendly architectures makes it an ideal candidate for 3D shape generation and reconstruction tasks. Hence, we consider the point cloud as our representation format.

In this work, we seek to answer three important questions in the tasks of semantic object reconstruction and segmentation:
\begin{enumerate*}[label=\textbf{(\arabic*)}]
    \item What is an effective way of inferring an accurate semantically annotated 3D point cloud representation of an object when provided with its two-dimensional image counterpart?
    \item How do we incorporate object geometry into the segmentation framework so as to improve segmentation accuracy?
    \item How do we incorporate semantic understanding into the reconstruction framework so as to improve the reconstruction of individual parts?
\end{enumerate*}
We achieve the former by training a neural network to jointly optimize for the reconstruction as well as segmentation losses. We empirically show that such joint training achieves superior performance on both reconstruction and segmentation tasks when compared to two different neural networks that are trained on each task independently. To enable the flow of information between the two tasks, we propose a novel loss formulation to integrate the knowledge from both the predicted semantics and the reconstructed geometry.

In summary, our contributions in this work are as follows:
\begin{itemize}
    \item We propose 3D-PSRNet, a part segmented 3D reconstruction network, which is jointly optimized for the tasks of reconstruction and segmentation.
    \item To enable the flow of information from one task to another, we introduce a novel loss function called \textit{location-aware segmentation loss}. We empirically show that the proposed loss function aids in the generation of more faithful part reconstructions, while also resulting in more accurate segmentations.
    \item We evaluate 3D-PSRNet on a synthetic dataset to achieve state-of-the-art performance in the task of semantic 3D object reconstruction from a single image. 
\end{itemize}

\section{Related Work}
\label{sec:related_work}

\noindent
\textbf{3D Reconstruction}

In recent times, deep learning based approaches have achieved significant progress in the field of 3D reconstruction. The earlier works focused on voxel-based representations ~\cite{girdhar2016learning,wu2016learning,choy20163d}. Girdhar \textit{et al.}~\cite{girdhar2016learning} map the 3D model and the corresponding 2D representations to a common embedding space to obtain a representation which is both predictable from 2D images and is capable of generating 3D objects. Wu \textit{et al.}~\cite{wu2016learning} utilize variational auto-encoders with an additional adversarial criterion to obtain improved reconstructions. Choy \textit{et al.}~\cite{choy20163d} employ a 3D recurrent network to obtain reconstructions from multiple input images. While the above works directly utilize the ground truth 3D models in the training stage, ~\cite{yan2016perspective,tulsiani2017multi,wu2017marrnet,zhu2017rethinking} try to reconstruct the 3D object using 2D observations from multiple view-points.     

Several recent works have made use of point clouds in place of voxels to represent 3D objects~\cite{fan2017point,groueix2018,mandikal20183dlmnet}. Fan \textit{et al.}~\cite{fan2017point} showed that point cloud prediction is not only computationally efficient but also outperforms voxel-based reconstruction approaches. Groueix \textit{et al.}~\cite{groueix2018} represented a 3D shape as a collection of parametric surface elements and constructed a mesh from the predicted point cloud. Mandikal \textit{et al.}~\cite{mandikal20183dlmnet} trained an image encoder in the latent space of a point cloud auto-encoder, while also enforcing a constraint to obtain diverse reconstructions. However, all of the above works focus solely on the point cloud reconstruction task.

\medskip
\noindent
\textbf{3D Semantic Segmentation}

Semantic segmentation using neural networks has been extensively studied in the 2D domain~\cite{long2015fully,he2017mask}. The corresponding task in 3D has been recently explored by works such as~\cite{song2017semantic,qi2017pointnet,qi2017pointnet++,kalogerakis20173d,muralikrishnan2018tags2parts}. Song \textit{et al.}~\cite{song2017semantic} take in a depth map of a scene as input and predict a voxelized occupancy grid containing semantic labels on a per-voxel basis. They optimize for the multi-class segmentation loss and argue that scene completion aids semantic label prediction and vice versa. Our representation format is a 3D point cloud while ~\cite{song2017semantic} outputs voxels. This gives rise to a number of differences in the training procedure. Voxel based methods predict an occupancy grid and hence optimize for the cross-entropy loss for both reconstruction as well as segmentation. On the other hand, point cloud based works optimize distance-based metrics for reconstruction and cross-entropy for segmentation. We introduce a location-aware segmentation loss tailored for point cloud representations.

~\cite{qi2017pointnet,qi2017pointnet++} introduce networks that take in point cloud data so as to perform classification and segmentation. They introduce network architectures and loss formulations that are are able to handle the inherent un-orderedness of the point cloud data.
While ~\cite{fan2017point} predicts only the 3D point cloud geometry from 2D images, and ~\cite{qi2017pointnet,qi2017pointnet++} segment input point clouds, our approach stresses the importance of jointly optimizing for reconstruction and segmentation while transitioning from 2D to 3D.

\section{Approach}
\label{sec:approach}

\begin{figure*}[!htb]
\centering
\begin{center}
\includegraphics[width=\linewidth]{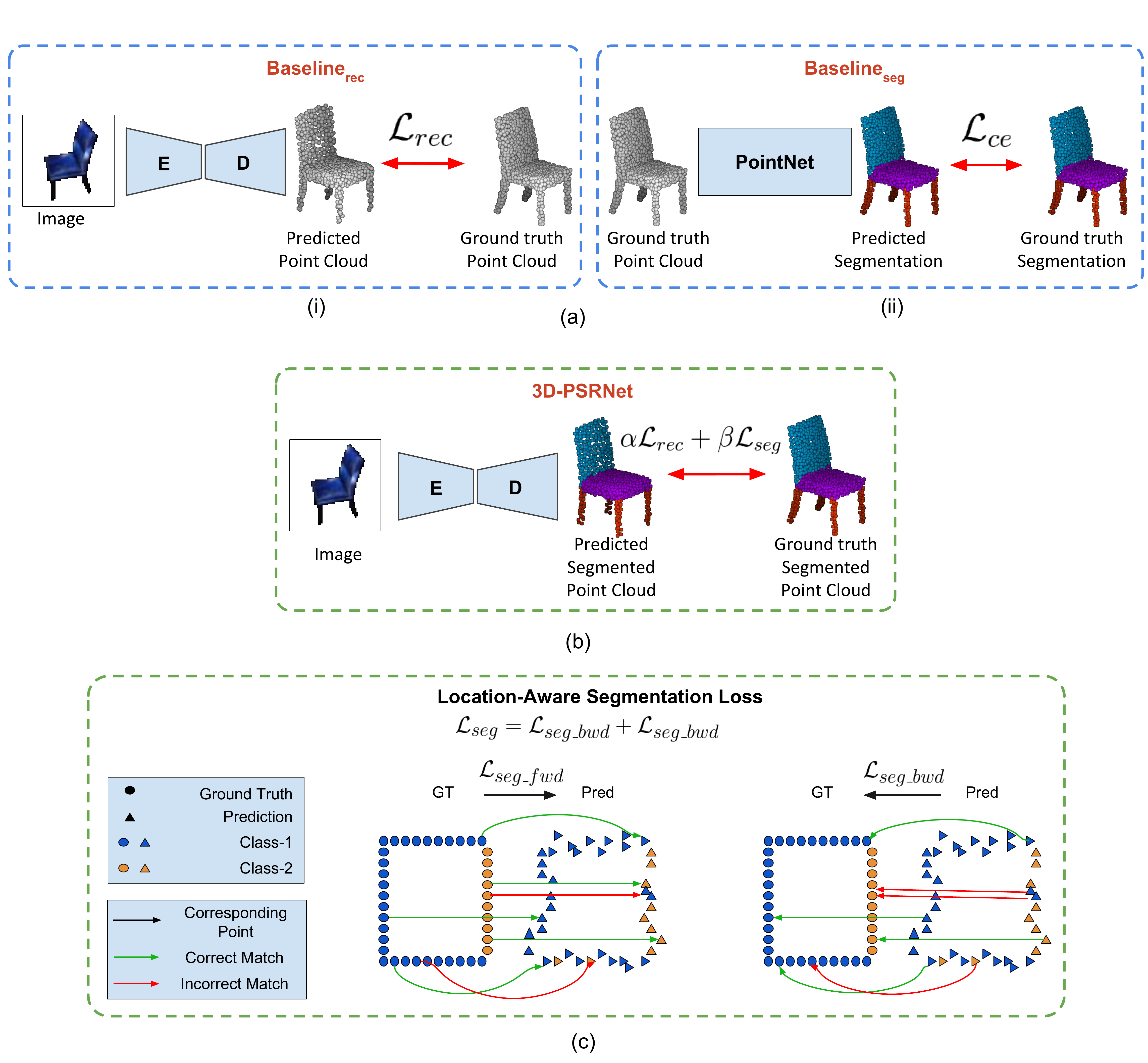}
\end{center}
\caption{Semantic point cloud reconstruction approaches. (a) Baseline: (i) A reconstruction network takes in an image input and predicts a 3D point cloud reconstruction of it. (ii) A segmentation network takes in a 3D point cloud as input and predicts semantic labels for every input point. (b) Our approach takes in an image as input and predicts a part segmented 3D point cloud by jointly optimizing for both reconstruction and segmentation, while also additionally propagating information from the semantic labels to improve reconstruction. (c) Point correspondences for location-aware segmentation loss. Incorrect reconstructions and segmentations are both penalized. The overall segmentation loss is the summation of the forward and backward segmentation losses.}
\label{fig:approach}
\end{figure*}

In this section, we introduce our model, 3D-PSRNet, which generates a part-segmented 3D point cloud from a 2D RGB image. As a baseline for comparison, we train two separate networks for the task of reconstruction and segmentation (Figure~\ref{fig:approach}(a)). Given an RGB image $I$ as input, the reconstruction network (\textit{baseline\textsubscript{rec}}) outputs a 3D point cloud $\widehat{X}_p\in\mathbb{R}^{{N}_p\times 3}$, where $N_p$ is the number of points in the point cloud. Given a 3D point cloud $X_p\in\mathbb{R}^{{N}_p\times 3}$ as input, the segmentation network (\textit{baseline\textsubscript{seg}}) predicts the class labels $\widehat{X}_{c}\in\mathbb{R}^{N_p\times N_c}$, where $N_c$ is the number of classes present in the object category. During inference, image $I$ is passed through \textit{baseline\textsubscript{rec}} to obtain $\widehat{X}_p$, which is then passed through \textit{baseline\textsubscript{seg}} to obtain $\widehat{X}_c$.

Our training pipeline consists of jointly predicting $(\widehat{X}_p$, $\widehat{X}_c)$ (Figure~\ref{fig:approach}(b)).
The reconstruction network is modified such that an additional $N_c$ predictions, representing the class probabilities of each point, are made at the final layer. The network is simultaneously trained with reconstruction and segmentation losses, as explained below.

\subsection{Loss Formulation}
\label{subsec:loss_formulation}
\noindent
\textbf{Reconstruction Loss}
We require a loss formulation that is invariant to the order of points in the point cloud. To satisfy this criterion, the Chamfer distance between the ground truth point cloud $X_p$ and predicted point cloud $\widehat{X}_p$ is chosen as the reconstruction loss. The loss function is defined as:  

\begin{equation}
    \mathcal{L}_{rec} = d_{Chamfer}(X_p,\widehat{X}_p) = \sum_{i\in X_p}\min_{j\in \widehat{X}_p}{||i-j||}^2_2 + \sum_{i\in \widehat{X}_p}\min_{j\in X_p}{||i-j||}^2_2
\label{eq:chamfer}
\end{equation}

\noindent
\textbf{Segmentation Loss}
We use point-wise softmax cross-entropy loss (denoted by $\mathcal{L}_{ce}$) between the ground truth class labels $X_c$ and the predicted class labels $\widehat{X}_c$. For the training of \textit{baseline\textsubscript{seg}}, since there is direct point-to-point correspondence between $X_p$ and $\widehat{X}_c$, we directly apply the segmentation loss as the cross-entropy loss between $X_c$ and $\widehat{X}_c$:
\begin{equation}
    \mathcal{L}_{ce}(X_c,\widehat{X}_c) = \sum_{\substack{x\in X_c \\ \hat{x}\in\widehat{X}_c}} [x \log(\hat{x}) + (1-x)\log(1-\hat{x})]
\end{equation}
However, during joint training, there exists no such point-to-point correspondence between the ground truth and predicted class labels. We therefore introduce the \textit{location-aware segmentation loss} to propagate semantic information between matching point pairs (Figure~\ref{fig:approach}(c)). The loss consists of two terms:
\begin{enumerate}[label=\textbf{(\arabic*)}]
    \item \textbf{Forward segmentation loss ($\mathcal{L}_{seg\_fwd}$)}: For every point $i\in X_p$, we find the closest point $i'\in \widehat{X}_p$, and apply $L_{ce}$ on their corresponding class labels.
        \begin{equation}
            \mathcal{L}_{seg\_fwd} = \frac{1}{N_p}\sum_{i\in X_p}\mathcal{L}_{ce}(X_{c_i}, \widehat{X}_{c_{i'}})
        \label{eq:L_seg}
        \end{equation}
        
    \item \textbf{Backward segmentation loss ($\mathcal{L}_{seg\_bwd}$)}: For every point $i\in \widehat{X}_p$, we find the closest point $i'\in {X}_p$, and apply $L_{ce}$ on their corresponding class labels.
        \begin{equation}
            \mathcal{L}_{seg\_bwd} = \frac{1}{N_p}\sum_{i\in \widehat{X}_p}\mathcal{L}_{ce}(X_{c_{i'}}, \widehat{X}_{c_i})
        \label{eq:L_seg}
        \end{equation}
        
\end{enumerate}

The overall segmentation loss is then the summation of the forward and backward segmentation losses:
\begin{equation}
    \mathcal{L}_{seg} = \mathcal{L}_{seg\_fwd} + \mathcal{L}_{seg\_bwd}
\label{eq:L_seg}
\end{equation}

The total loss during joint training is then given by,
\begin{equation}
    \mathcal{L}_{tot} = \alpha \mathcal{L}_{rec} + \beta \mathcal{L}_{seg}
\label{eq:L_tot}
\end{equation}

\subsection{Implementation Details}
For training the baseline segmentation network \textit{baseline\textsubscript{seg}}, we follow the architecture of the segmentation network of PointNet~\cite{qi2017pointnet}, which consists of ten 1D convolutional layers of filter sizes $[64,64,64,128,1024,512,256,128,128,N_c]$, where $N_c$ is the number of class labels. A global maxpool function is applied after the fifth layer and the resulting feature is concatenated with each individual point feature, as is done in the original paper. Learning rate is set to $5e^{-4}$ and batch normalization is applied at all the layers of the network. 
The networks for the baseline reconstruction network and the joint 3D-PSRNet are similar in architecture. They consist of four 2D convolutional layers with number of filters as $[32,64,128,256]$, followed by four fully connected layers with output dimensions of size $[128,128,128,N_p\times 3]$ (reconstruction) and $[128,128,128,N_p\times (3+N_c)]$ (joint), where $N_p$ is the number of points in the point cloud. We set $N_p$ to be 1024 in all our experiments. Learning rate for \textit{baseline\textsubscript{rec}} and 3D-PSRNet are set to $5e^{-5}$ and $5e^{-4}$ respectively. We use a minibatch size of 32 in all the experiments. We train the individual reconstruction and segmentation networks for 1000 epochs, while the joint network (3D-PSRNet) is trained for 500 epochs. We choose the best model according to the corresponding minimum loss. In Eq.~\ref{eq:L_tot}, the values of $\alpha$ and $\beta$ are set to $1e^4$ and $1$ respectively for joint training. Codes are available at \url{https://github.com/val-iisc/3d-psrnet}.


\section{Experiments}
\label{sec:evaluation}

\subsection{Dataset}

We train all our networks on synthetic models from the ShapeNet dataset~\cite{chang2015shapenet} whose part annotated ground truth point clouds are provided by~\cite{Yi16}. Our dataset comprises of 7346 models from three exemplar categories - chair, car and airplane. We render each model from ten different viewing angles with azimuth values in the range of $[0,360]$ and elevation values in the range of $[-20,40]$ so as to obtain a dataset of size 73,460. We use the train/validation/test split provided by~\cite{Yi16} and train a single model on all the categories in all our experiments.

\subsection{Evaluation Methodology}

\begin{enumerate}[label=\textbf{(\arabic*)}]
    \item \textbf{Reconstruction}: We report both the Chamfer Distance (Eqn.~\ref{eq:chamfer}) as well as the Earth Mover's Distance (or EMD) computed on 1024 points in all our evaluations. EMD between two point sets $\widehat{X}_{\!_P}$ and $X_{\!_P}$ is given by:
        \begin{equation}
        \label{eq:emd}
            d_{EMD}(X_p,\widehat{X}_p)=\min_{\phi:X_p\rightarrow \widehat{X}_p}\sum_{x\in X_p}||x-\phi(x)||_2
        \end{equation}
    where $\phi:X_p\rightarrow \widehat{X}_p$ is a bijection. For computing the metrics, we renormalize both the ground truth and predicted point clouds within a bounding box of length 1 unit.
    
    \medskip
    \item \textbf{Segmentation}: We formulate part segmentation as a per-point classification problem. Evaluation metric is mIoU on points. For each shape S of category C, we calculate the shape mIoU as follows: For each part type in category C, compute IoU between groundtruth and prediction. If the union of groundtruth and prediction points is empty, then count part IoU as 1. Then we average IoUs for all part types in category C to get mIoU for that shape. To calculate mIoU for the category, we take average of mIoUs for all shapes in that category. Since there is no correspondence between the ground truth and predicted points, we use a mechanism similar to the one described in Section~\ref{subsec:loss_formulation} for computing the forward and backward mIoUs, before averaging them out to get the final mIoU as follows:
    
    \begin{equation}
    \label{eq:miou}    
    \begin{aligned}{}
        mIoU(X_c, \widehat{X}_c) = & \frac{1}{2N_c}\sum_{i}\frac{N_{ii}}{\sum_{j}N_{ij}+\sum_{j}N_{ji}-N_{ii}} \\ & + \frac{1}{2N_c}\sum_{i}\frac{\widehat{N}_{ii}}{\sum_{j}\widehat{N}_{ij}+\sum_{j}\widehat{N}_{ji}-\widehat{N}_{ii}}
    \end{aligned}
    \end{equation}
    where $N_{ij}$ is the number of points in category $i$ in $X_c$ predicted as category $j$ in $\widehat{X}_c$ for forward point correspondences between $X_c$ and $\widehat{X}_c$. Similarly $\widehat{N}_{ij}$ is for backward point correspondences. $N_c$ is the total number of categories.
\end{enumerate}

\begin{table}[t]
\setlength{\tabcolsep}{7pt}
\centering
\begin{center}
\begin{tabular}{cccc}
\toprule
Category                  & Metric  & \begin{tabular}[c]{@{}c@{}}PSGN-FC~\cite{fan2017point}\\ + PointNet~\cite{qi2017pointnet}\end{tabular} & 3D-PSRNet\\
\noalign{\smallskip}
\hline
\noalign{\smallskip}
\multirow{3}{*}{Chair}    & Chamfer & 6.82    & \textbf{6.57}  \\
                          & EMD     & 11.37    & \textbf{10.10}        \\
                          & mIoU    & 78.09   & \textbf{81.92}         \\
                          \noalign{\smallskip}
                          \hline
                          \noalign{\smallskip}
\multirow{3}{*}{Car}      & Chamfer & 5.48    & \textbf{5.14}         \\
                          & EMD     & 5.88    & \textbf{5.53}         \\
                          & mIoU    & 59.0    & \textbf{61.57}         \\
                          \noalign{\smallskip}
                          \hline
                          \noalign{\smallskip}
\multirow{3}{*}{Airplane} & Chamfer & \textbf{4.06}        & \textbf{4.06}         \\
                          & EMD     & 7.06        & \textbf{6.24}         \\
                          & mIoU    & 62.86       & \textbf{68.64}         \\
                          \bottomrule
                          \noalign{\smallskip}
\multirow{3}{*}{\textbf{Mean}}  & Chamfer & 5.45   & \textbf{5.26}   \\
                          & EMD     & 8.10         & \textbf{7.29}         \\
                          & mIoU    & 66.65        & \textbf{70.71}         \\
                          \bottomrule
\end{tabular}
\end{center}
\caption{Reconstruction and Segmentation metrics on ShapeNet~\cite{chang2015shapenet}. 3D-PSRNet significantly outperforms the baseline in both the reconstruction and segmentation metrics on all categories. Chamfer and EMD metrics are scaled by 100.}
\label{tab:sota_shapenet}
\end{table}

\begin{figure*}[h!]
\centering
\begin{center}
    \includegraphics[width=0.85\linewidth]{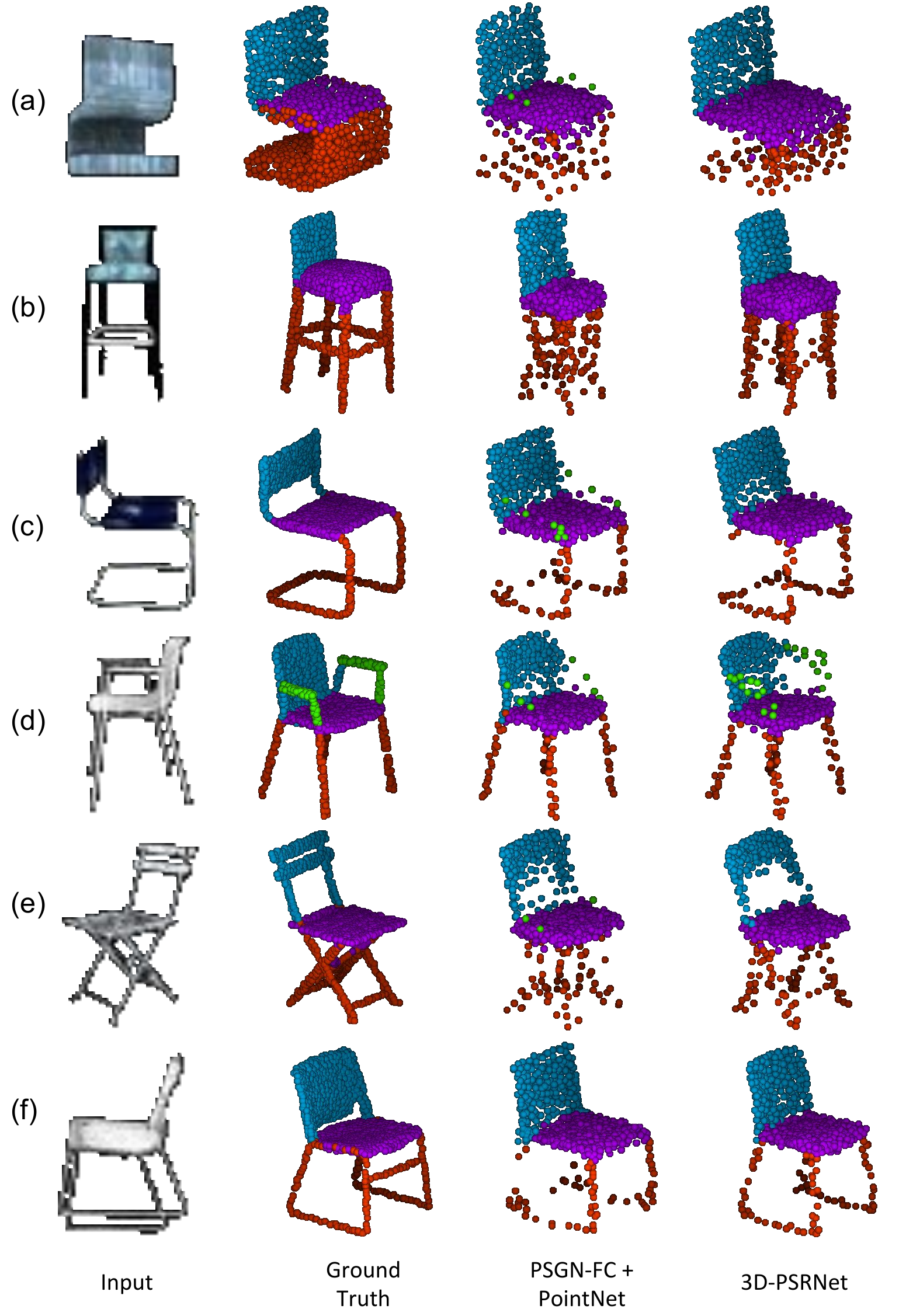}
\end{center}
\caption{Qualitative results on the chair category from ShapeNet~\cite{chang2015shapenet}. Compared to the baseline (PSGN~\cite{fan2017point} + PointNet~\cite{qi2017pointnet}), we are better able to capture the details present in the input image. Individual parts such as legs (b,e,f) and handles (d) are reconstructed with greater accuracy. Additionally, while outlier points are present in the baseline (a,c), our method produces more uniformly distributed reconstructions.}
\label{fig:shapenet_comparison1}
\end{figure*}

\begin{figure*}
\centering
\begin{center}
    \includegraphics[width=\linewidth]{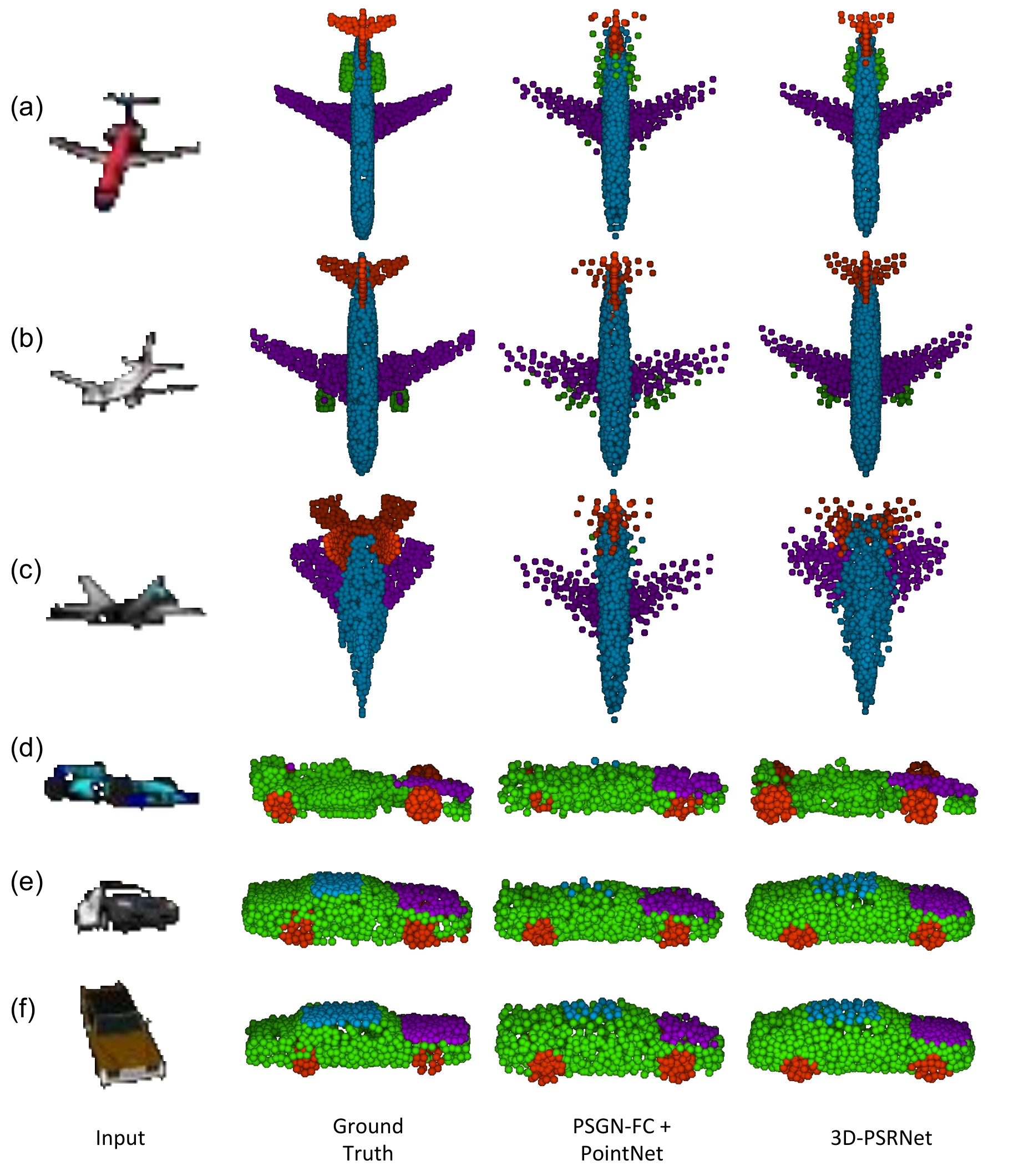}
\end{center}
\caption{Qualitative results on airplanes and cars from ShapeNet~\cite{chang2015shapenet}. Compared to the baseline (PSGN~\cite{fan2017point} + PointNet~\cite{qi2017pointnet}), we are better able to reconstruct individual parts in each category resulting in better overall shape. Our method produces sharper reconstruction of tails and wings in airplanes (a,b). We also obtain more uniformly distributed points (as is visible in the wing region of airplanes). In cars, our reconstructions better correspond to the input image compared to the baseline.}
\label{fig:shapenet_comparison2}
\end{figure*}

\subsection{Results}

Table~\ref{tab:sota_shapenet} presents the quantitative results on ShapeNet for the baseline and joint training approaches. 3D-PSRNet achieves considerable improvement in both the reconstruction (Chamfer, EMD) and segmentation (mIoU) metrics. It outperforms the baseline approach in every metric on all categories. On an average, we obtain \textbf{4.1\%} improvement in mIoU. 

\begin{figure*}[h!]
\centering
\begin{center}
    \includegraphics[width=0.75\linewidth]{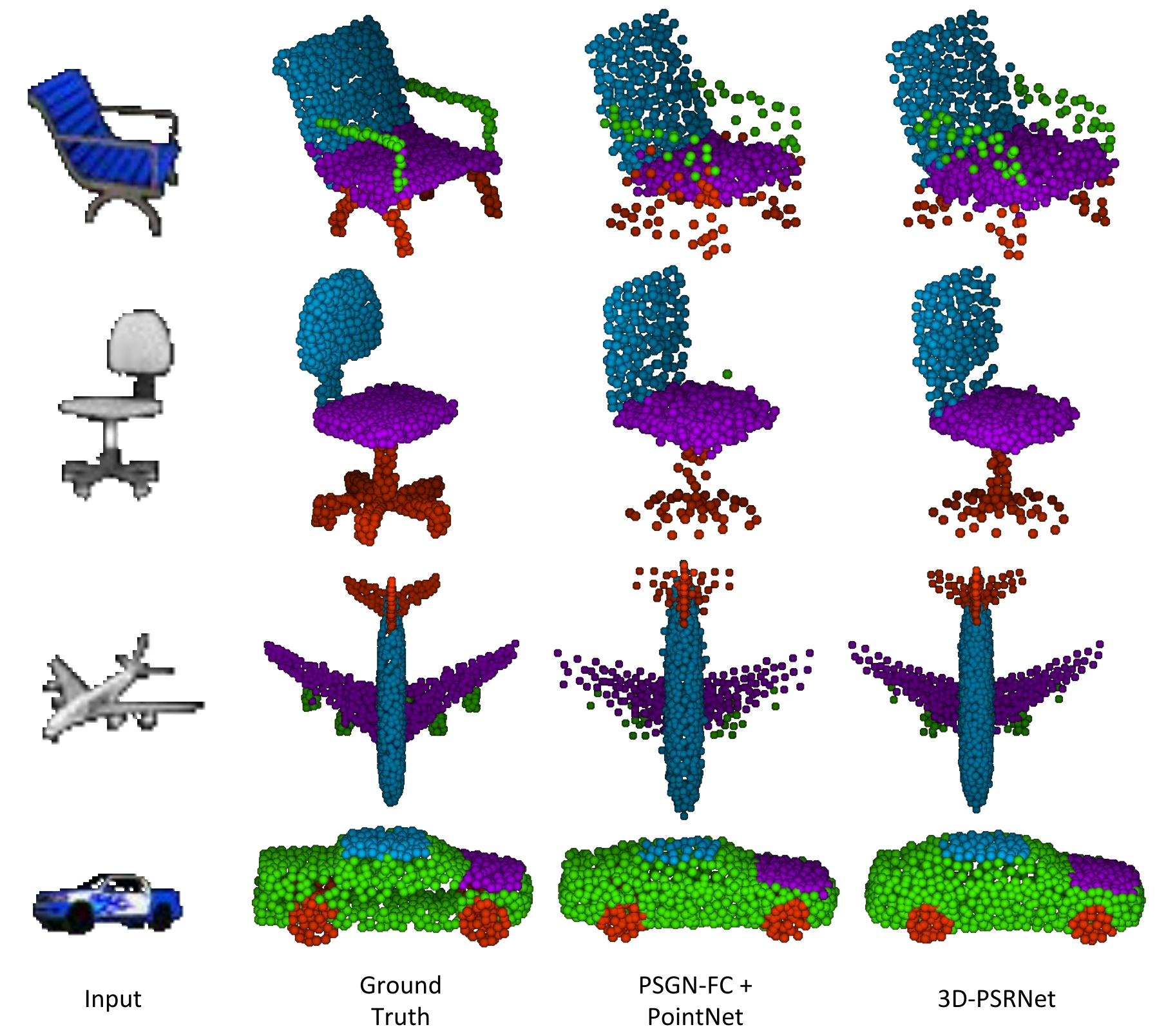}
\end{center}
\caption{Failure cases of our method. We notice that our method fails to get finer details in some instances, such as leg details in chairs, dual turbines present in airplanes, and certain car types.}
\label{fig:shapenet_failure}
\end{figure*}

The qualitative results are presented in Figures~\ref{fig:shapenet_comparison1} and~\ref{fig:shapenet_comparison2}. 3D-PSRNet obtains more faithful reconstructions compared to the baseline to achieve better correspondence with the input image. It also predicts more uniformly distributed point clouds. We observe that joint training results in reduced hallucination of parts (for e.g. predicting handles for chairs without handles) and spurious segmentations. We also show a few failure cases of our approach in Figure~\ref{fig:shapenet_failure}. The network misses out on some finer structures present in the object (e.g. dual turbines in the case of airplanes). The reconstructions are poorer for uncommon input samples. However, these drawbacks also exist in the baseline approach.    

\vspace{-6pt}
\subsection{Relative Importance of Reconstruction and Segmentation losses}

\begin{figure*}[h!]
    \centering
    \begin{minipage}{\textwidth}
        \begin{minipage}{0.32\textwidth}
            \centering
            \includegraphics[scale=0.2]{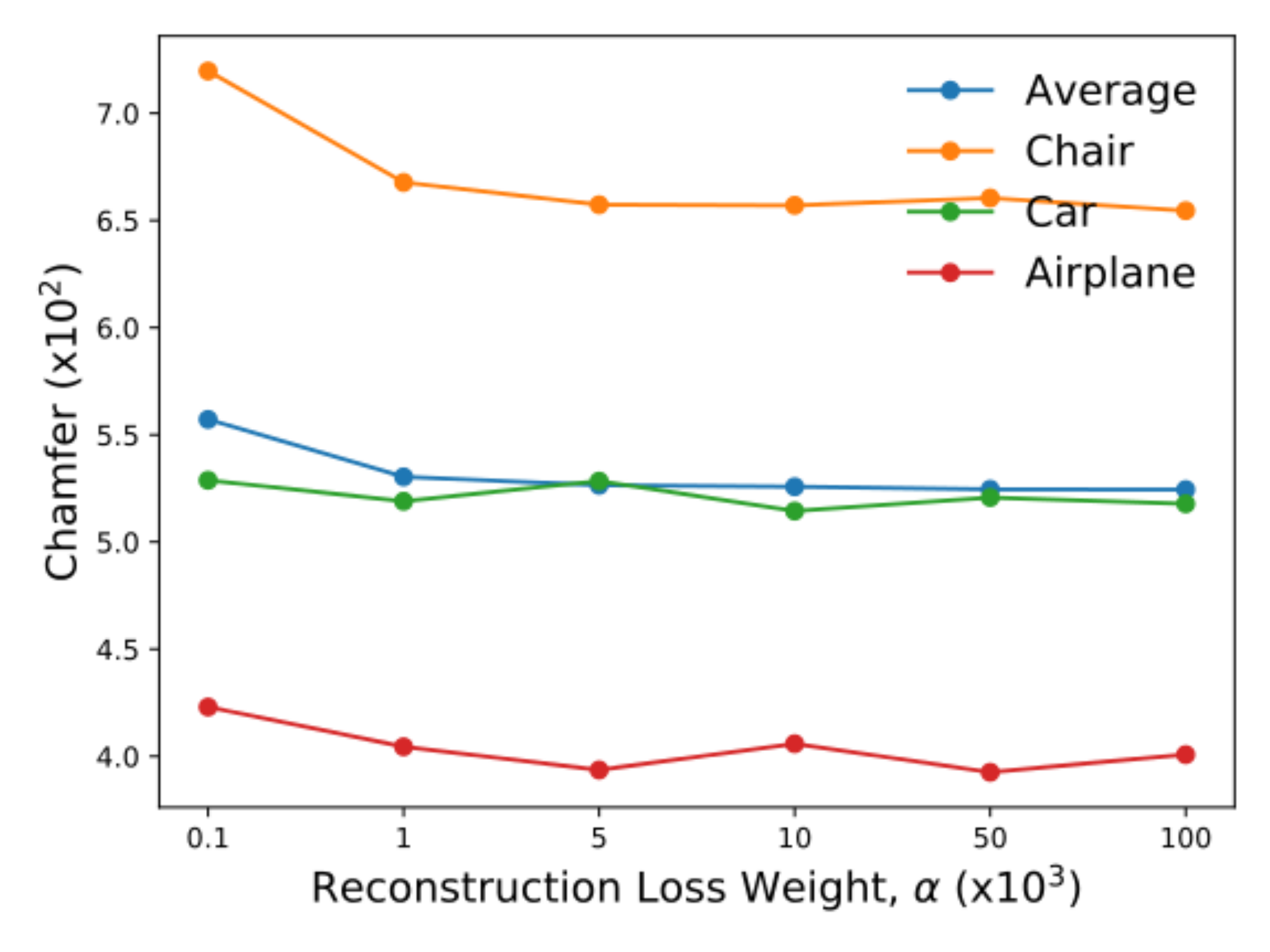}
        \end{minipage}
        \begin{minipage}{0.32\textwidth}
            \centering
            \includegraphics[scale=0.2]{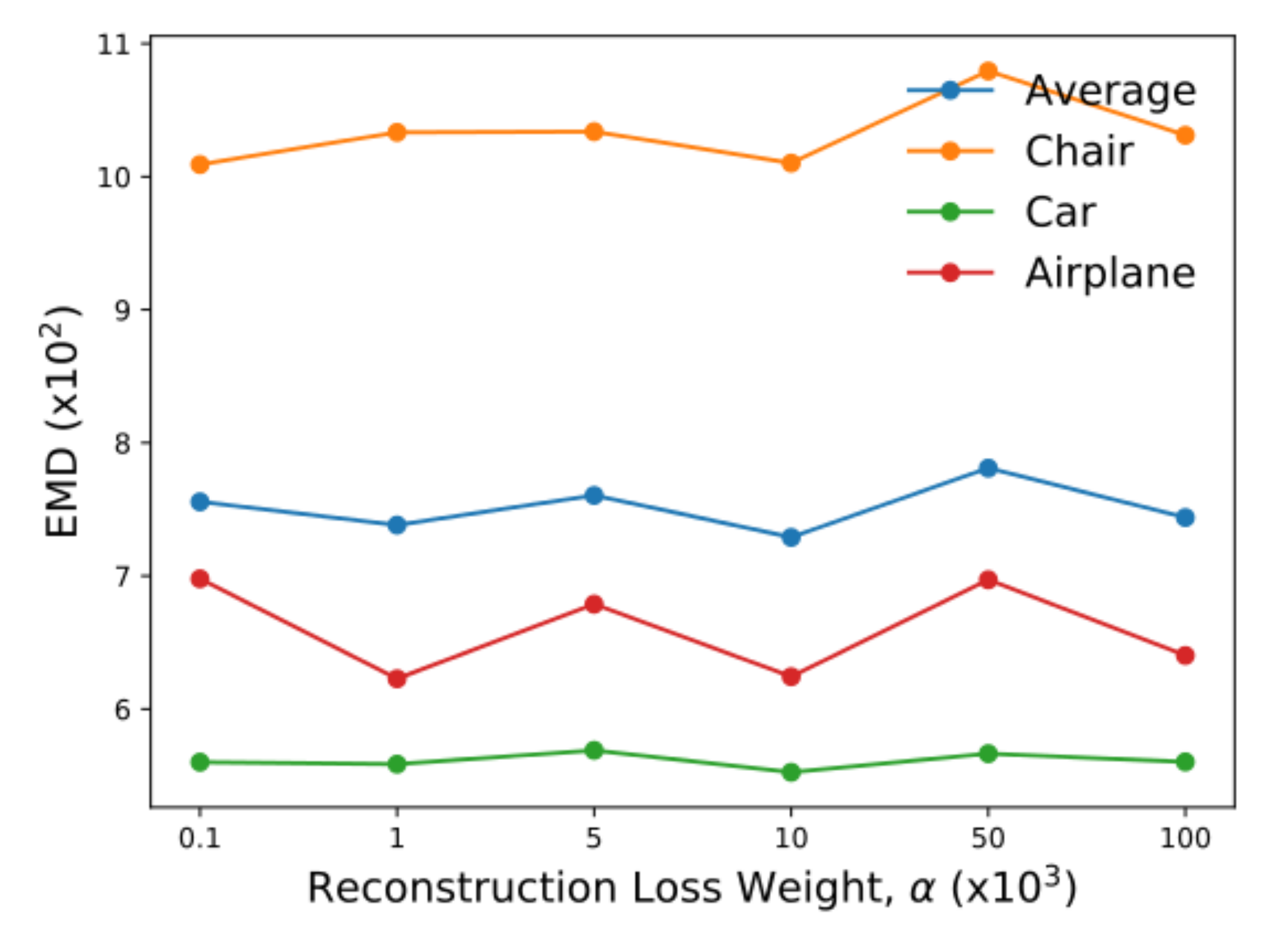}
        \end{minipage}
        \begin{minipage}{0.32\textwidth}
            \centering
            \includegraphics[scale=0.2]{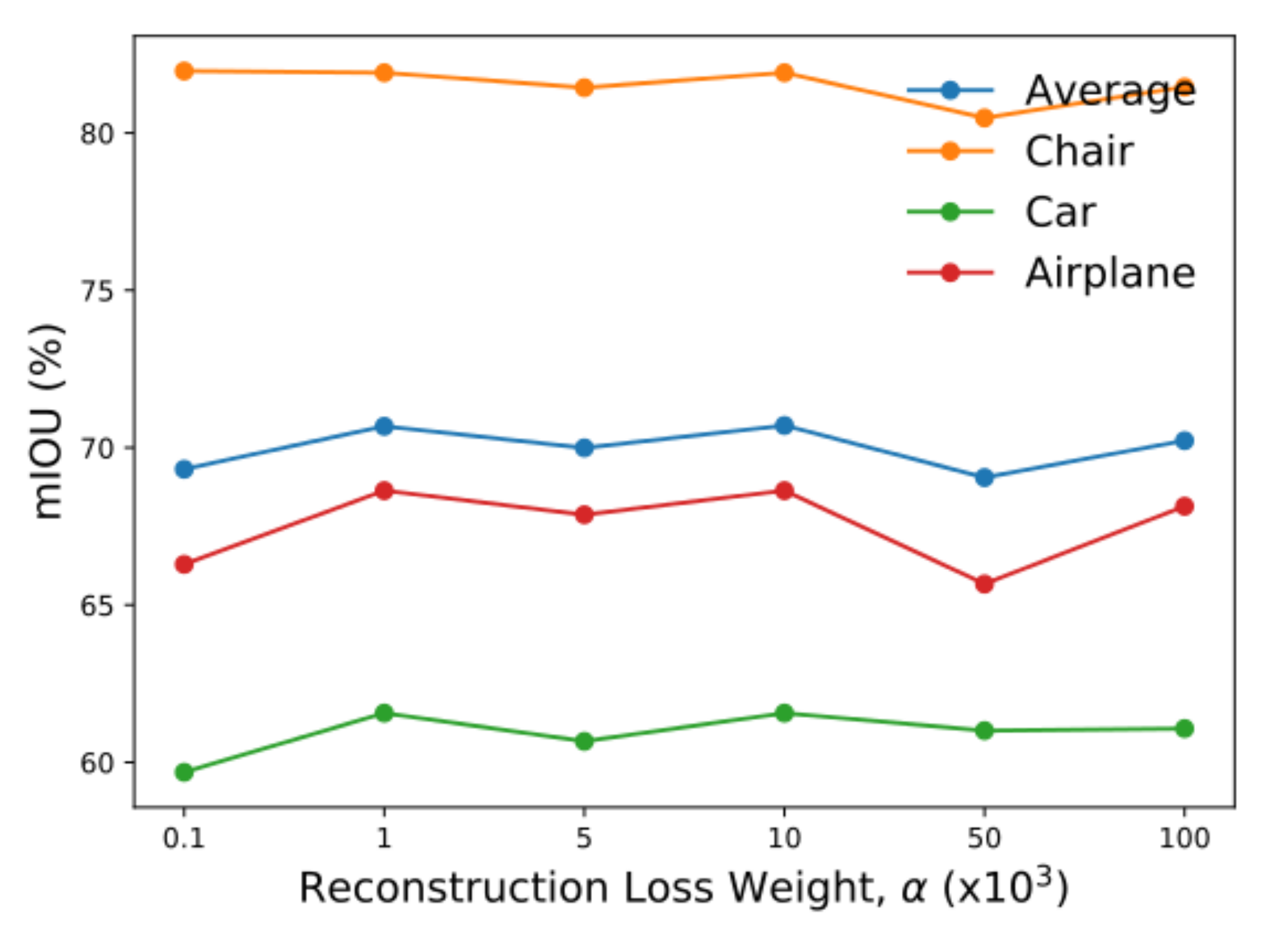}
        \end{minipage}
    \end{minipage}
    \caption{Ablative study on weight for reconstruction loss, $\alpha$. Chamfer, EMD and mIOU metrics are calculated for different values of $\alpha$. Based on the plots, we choose the value of $\alpha$ to be $10^4$.}
    \label{fig:rec_loss_alpha}
\end{figure*}

We present an ablative study on the relative weightage of the reconstruction and segmentation losses in Eq.~\ref{eq:L_tot}. We fix the value of $\beta$ to one, while $\alpha$ is varied from $10^2$ to $10^5$. Figure~\ref{fig:rec_loss_alpha} presents the plot of Chamfer, EMD and mIoU metrics for varying values of $\alpha$. We observe that for very low value of $\alpha$, both the reconstruction and segmentation metrics are worse off, while there is minimal effect on the average metrics for $\alpha$ greater than $10^3$. Based on Figure~\ref{fig:rec_loss_alpha}, we set the value of $\alpha$ to $10^4$ in all our experiments.

\section{Conclusion}
\label{sec:conclusion}

In this paper, we highlighted the importance of jointly learning the tasks of 3D reconstruction and object part segmentation. We introduced a loss formulation in the training regime to enable propagating information between the two tasks so as to generate more faithful part reconstructions while also improving segmentation accuracy. We thoroughly evaluated against existing reconstruction and segmentation baselines, to demonstrate the superiority of the proposed approach. Quantitative and qualitative evaluation on the ShapeNet dataset demonstrate the effectiveness in generating more accurate point clouds with detailed part information in comparison to the current state-of-the-art reconstruction and segmentation networks.

\bibliographystyle{splncs04}
\bibliography{egbib}
\end{document}